\documentclass{article}
\usepackage[preprint,nonatbib]{neurips_2018}

\usepackage{longtable}

\usepackage{booktabs}

\usepackage{mathrsfs}
\usepackage{algorithm}
\usepackage{algorithmic}
\usepackage{wrapfig}
\usepackage{xcolor}

\usepackage{multirow}
\usepackage{lscape}
\usepackage{tabulary}
\usepackage{bbm}

\usepackage[utf8]{inputenc} 
\usepackage[T1]{fontenc}    
\usepackage{hyperref}       
\usepackage{url}            
\usepackage{booktabs}       
\usepackage{amsfonts}       
\usepackage{nicefrac}       
\usepackage{microtype}      

\usepackage{amsmath}
\usepackage{algorithm, algorithmic}
\usepackage{amsfonts, amssymb}
\usepackage{multirow}
\usepackage{mathtools}
\usepackage{arydshln}
\usepackage{refcount}
\usepackage{listings}
\usepackage{hyperref}

\newtheorem{theorem}{Theorem}[section]

\newtheorem{proof}[theorem]{Proof}
\newtheorem{proposition}[theorem]{Proposition}

\DeclareMathOperator*{\argmin}{arg\,min}

\title{Regularize! Don't Mix: Multi-Agent Reinforcement Learning without Explicit Centralized Structures }

\author{%
  Chapman Siu \\
  Faculty of Engineering and Information Technology \\
  University of Technology Sydney, Australia \\
  \texttt{chpmn.siu@gmail.com} \\
  \And
   Jason Traish \\
  Faculty of Engineering and Information Technology \\
  University of Technology Sydney, Australia \\
  \texttt{Jason.Traish@uts.edu.au} \\
  \AND
  Richard Yi Da Xu \\
  Faculty of Engineering and Information Technology \\
  University of Technology Sydney, Australia \\
  \texttt{YiDa.Xu@uts.edu.au} \\
}

\begin{document}

\maketitle

\begin{abstract}
We propose using regularization for Multi-Agent Reinforcement Learning rather than learning explicit cooperative structures called {\em Multi-Agent Regularized Q-learning} (MARQ). Many MARL approaches leverage centralized structures in order to exploit global state information or removing communication constraints when the agents act in a decentralized manner. Instead of learning redundant structures which is removed during agent execution, we propose instead to leverage shared experiences of the agents to regularize the individual policies in order to promote structured exploration. We examine several different approaches to how MARQ can either explicitly or implicitly regularize our policies in a multi-agent setting. MARQ aims to address these limitations in the MARL context through applying regularization constraints which can correct bias in off-policy out-of-distribution agent experiences and promote diverse exploration. Our algorithm is evaluated on several benchmark multi-agent environments and we show that MARQ consistently outperforms several baselines and state-of-the-art algorithms; learning in fewer steps and converging to higher returns. 
\end{abstract}

\section{Introduction}


Reinforcement learning (RL) has increasingly greater real-world applications, ranging from autonomous vehicle coordination \cite{autocar}, and controlling robotic swarms \cite{guided} which can be framed as multi-agent cooperative problems. In these settings, agents act only on their partial observability and/or communication constraints leading to learning {\em decentralized policies}, which only rely on each agent's local state-action history. Although single-agent RL algorithms can be applied to these environments \cite{iql}, this approach is not inherently scalable nor does it allow for agents to leverage each others experience or encourage structured exploration in the cooperative multi-agent setting.

In Multi-Agent Reinforcement Learning (MARL), the nature of the joint state action space leads to complexity in estimation, as the size of the joint action spaces grow exponentially with the number of agents. Leveraging traditional Reinforcement Learning approaches which model the combined joint action and state space is often impractical or does not appropriate reflect constraints which may be present in the evaluation environment such as partial observability and/or communication constraints leading to learning {\em decentralized policies}. Attempting to leverage experience of other agents in Q-Learning set up does not greatly improve performance as off-policy samples would not have been corrected from the underlying policy distribution \cite{seac}, whereas using maximum entropy RL on an individual agent may not guarantee diverse exploration across all agents, requiring a network to learn the appropriate credit assignment \cite{lica}. Structured explorations have been explored through to explicitly learning a latent policy conditional on global state information \cite{maven}. MARQ aims to address these limitations in the MARL context without explicit cooperative structures through applying regularization constraints which can correct bias in off-policy out-of-distribution agent experiences and promote diverse exploration. 

To achieve this, rather than attempting to model and learn cooperative structures \cite{qmix,lica,maddpg,maven}, we instead treat the MARL challenge as an independent reinforcement learning problem \cite{iql} and examine different approaches to how we can leverage information from other agents in the MARL setting. Our approach involves encouraging more structured exploration through regularization the gradient updates across different agents. In effect, this could allow agents to search in orthogonal directions to encourage the maximum coverage of state-action combinations or facilitate in using their collective experience to determine the most promising approaches. In our experiments we demonstrate the viability of these approaches which can lead to greater returns in less iterations.

We demonstrate these ideas in a Q-learning framework, which we call {\em Multi-Agent Regularized Q-learning} (MARQ), which operates on independent learning \cite{iql} and leverages distributional reinforcement learning approach \cite{qrdqn}. Using a distributional approach allows for the distribution of returns to be modeled explicitly instead of estimating the mean. In estimating the distribution, we can begin to apply ideas including importance sampling and divergence measures to control how experiences are collected, and the direction in which policies gravitate towards to enable a more structured exploration approach leading to faster learner.

Although previous work leverage shared experiences in the actor critic framework and regularized policies in the off-policy setting, they are not directly applicable in Multi-Agent settings where additional considerations in the joint state-action space needs to be considered. Our primary contribution is introducing {\em Multi-Agent Regularized Q-learning} (MARQ), and can be summarized as follows.

\begin{itemize}
    \item MARQ regularizes the Q-values during training to encourage more structured exploration compared with existing Q-learning approaches, leading to more efficient training
    \item We demonstrate empirically our approach which builds on top of independent learning approaches across a range of benchmark MARL environments. MARQ outperforms other Q-learning approaches and independent learning approaches with minimal overhead. 
\end{itemize}

\section{Related Work}

\subsection*{Structured Exploration}

One area of interest is promoting diversification when exploring the environment. This has been approached through attempting to discover new states in an unsupervised manner \cite{diayn}, which trains on diversity of rewards and through encouraging action diversification in considering the joint state and action through mutual information in MARL setting \cite{maven}. Another way is to modify the rewards in order to improve the tradeoff between exploitation and exploration \cite{vime,curious}.

\subsection*{Entropy Regularization}

A related approach to structured exploration is entropy regularization. A common technique to improve policy optimization is through stochastic actions which is achieved through adding a entropy regularization penalty. Existing methods are typically used in single-agent RL which modify the training objective \cite{sac,softql,energyentropy}, other approaches include augmenting the reward itself \cite{mrl}. When used in the multi-agent setting it has been shown that an adaptive approach to regularization is required in order to keep consistent exploration across all agents to generate optimal joint actions in MARL environments \cite{lica}. In additional to entropy regularization, we provide explicit consideration to encourage diversity across the agents. 

\subsection*{Distributional Reinforcement Learning}

Recent approaches rely on distributional reinforcement learning in order to estimate the distribution of the value function in Q-learning approaches \cite{qrdqn}. These have allowed for implicit regularization of the policy through augmenting the reward \cite{mrl}, providing off-policy learning guarantees when leveraging external, offline experience gathering techniques \cite{cql}. We offer extensions to the distributional reinforcement learning approach in the multi-agent setting, with cooperative considerations.

\subsection*{Learning from Experience}  

Offline reinforcement learning \cite{cql} is a promising field which can handle distribution shifts due to experiences being sampled from an alternative policy which are not possible in standard single-agent RL \cite{offlinetutorial}, in which distributional RL has shown promise. In the multi-agent setting, shared experience for actor critic models has been tackled through correcting off-policy sampling through importance sampling \cite{seac}. Our approach leverages these corrections in the multi-agent Q-learning space which previous was explored with limited impact \cite{seac}.

\section{Preliminaries}


\subsection*{Markov Games}

The goal in reinforcement learning is to maximize the expected cumulative reward in a Markov decision process, through optimizing a learned policy. We define the tuple $\langle \mathcal{N}, G, \{S_{\mathsf{a}_i} \}_{\mathsf{a}_i \in \mathcal{N}}, \mathcal{A}, T, r, \gamma \rangle$ to represent partially observable Markov game for agents $\mathsf{a}_i \in \mathcal{N} = \{\mathsf{a}_1, \dots, \mathsf{a}_N\}$ \cite{markovgame}, global state space $G$, and joint action space $\mathcal{A} = A^1 \times \dots \times A^N$. Each agent $\mathsf{a}_i$ only perceives $s_{\mathsf{a}_i} \in S_{\mathsf{a}_i}$ which depends on the current global state, $T(s^\prime_{\mathsf{a}_i} \vert s_{\mathsf{a}_i}, a_{\mathsf{a}_i})$ and $r(s_{\mathsf{a}_i}, a_{\mathsf{a}_i}$ represent the dynamics and reward function, with $\gamma \in (0, 1)$ being the discount factor. Let $\langle s, a, r, s^\prime \rangle \sim \mathcal{D}_{\mathsf{a}_i}$ represent the dataset of experiences stored as a transition tuple for agent $\mathsf{a}_i$, with $s^\prime$ representing the state observed after taking action $a$ in state $s$ and receiving reward $r$. In this paper we are particularly interested in multi-agent algorithms which only use each agent's local observation during training and execution, meaning no global state information is used. 

\subsection*{Deep Q-Learning}

Q-learning is an off-policy RL algorithm based on dynamic programming, which uses the action-value function $Q(s, a)$.  In Deep learning approaches parameters this Q function uses a deep neural network \cite{dqn,qrdqn}. Q-learning methods apply the Bellman operator $\mathcal{B}^*Q(s, a) = r(s, a) + \gamma \mathbb{E}_{s^\prime \sim P(s^\prime \vert s, a)}[ \max_{a^\prime} Q(s^\prime, a^\prime)]$. Given the tuple $\langle s, a, r, s^\prime \rangle \sim \mathcal{D}_{\mathsf{a}_i}$, the iterative updating for training a particular agent $\mathsf{a}_i$ with tradeoff factor $\alpha$:

\begin{equation}\label{eq:dqn}
    {{\hat{Q}}^{k+1}} \leftarrow  \argmin_Q \alpha \cdot \mathbb{E}_{s \in {\mathcal{D}}, a \sim \pi(a \vert s)}[Q(s, a)]+ \frac{1}{2} \mathbb{E}_{s,a, r s^\prime  \sim \mathcal{D}}\left[(Q(s, a) - \mathcal{B}^*{\hat{Q}^{k}}(s, a) )^2\right]
\end{equation}

Where $\hat{Q}$ is typically a {\em target network} where the parameters kept constant for a number of iterations and periodically copied. 

\subsection*{Conservative Q-Learning}
\label{cqlr}

Conservative Q-Learning (CQL) extends approaches taken in DQN, which leverages information from the empirical behavioral distribution. Let $\hat{\pi}_{\mathcal{D}}(a \vert s) := \frac{\sum_{s, a \in \mathcal{D}} \mathbf{1}[s=s, a=a]}{\sum_{s \in \mathcal{D}} \mathbf{1}[s=s]}$ denote the empirical behavioral distribution at the state $s$. Then by adding an additional Q-value {\em maximization} term (in red) under the data $\hat{\pi}_{\mathcal{D}}$, we have the resulting iterative update for CQL:

\begin{align}
    {{\hat{Q}}^{k+1}} \leftarrow  \argmin_Q \alpha \cdot \Big(\mathbb{E}_{s \sim {\mathcal{D}}, a \sim \pi(a \vert s)}[Q(s, a)] {\color{red}- \mathbb{E}_{s \sim \mathcal{D}, a \sim \hat{\pi}(a \vert s)} [Q(s, a)]}\Big) \nonumber \\
    \label{eq:cql}
    + \frac{1}{2} \mathbb{E}_{s,a, r s^\prime  \sim \mathcal{D}}\left[(Q(s, a) - \mathcal{B}^*{\hat{Q}^{k}}(s, a) )^2\right]
\end{align}

This approach provides lower bound on the value of the current policy which provides theoretical improvement guarantees. In our work, we are interested in extensions of this approach through choices of regularizers $\mathcal{R}(\pi)$, in the multi-agent setting, when additional considerations are made with respective to leveraging other agent experiences or inducing a more structured exploration strategy. This is described in the form:

\begin{align}
    {{\hat{Q}}^{k+1}} \leftarrow  \argmin_Q \alpha \cdot \Big(\mathbb{E}_{s \sim {\mathcal{D}}, a \sim \pi(a \vert s)}[Q(s, a)] {- \mathbb{E}_{s \sim \mathcal{D}, a \sim \hat{\pi}(a \vert s)} [Q(s, a)]}\Big) \nonumber \\
    \label{eq:cqlr}
    + \frac{1}{2} \mathbb{E}_{s,a, r s^\prime  \sim \mathcal{D}}\left[(Q(s, a) - \mathcal{B}^*{\hat{Q}^{k}}(s, a) )^2 \right] + {\color{red} \mathcal{R}(\pi)}
\end{align}

CQL presents several variations for the regularizer $\mathcal{R}$ with closed form solutions, including entropy approach $\mathcal{R}(\pi) = H(\pi)$ and a KL-divergence approach to a prior distribution $\rho$ in the form $\mathcal{R}(\pi) = -D_{\text{KL}}(\pi \Vert \rho)$. These approaches have been empirically demonstrated to be more stable in high-dimensional action spaces \cite{cql}.

\section{Multi-Agent Regularized Q-Learning}

In this section, we develop {\em Multi-Agent Regularized Q-learning} (MARQ) approach, which employs independent Q-learning with allowances for shared experiences and structured exploration to promote cooperation. In the MARL setting independent Q-learning operates through having each agent learn its own policy, conditioned on its own partial observation \cite{iql}. We consider a {\em Conservative Q-Learning} framework leveraging distributional Q-learning for discrete actions in the form of Quantile Regression DQN, in which the distribution of returns is explicitly modelled rather than only the mean \cite{qrdqn}. The advantages of our approach is that it does not require learning centralized training frameworks such as QMIX, and can scale to an arbitrary number of agents, leading to faster training iterations. This is achieved through accounting for the distributional shift between the dataset and the learned policy, which is particularly importance when considering the large joint state space.









\subsection{Correcting for Shared Experiences}

\label{sharedexp}
One approach in the multi-agent setting is for Q-learning is to leverage other agents experience in the iterative update for the Q-function. It can be shown that using sampling other agent experiences indiscriminately can lead to underestimation of the Q function.  

\begin{proposition}

Suppose we are interested in updating the Q-function for agent $1$ using experiences from agent $2$, with the iterative update based on applying the DQN iterative update in Equation \ref{eq:dqn} with other agent experiences:

\begin{equation}\label{qiterate}
    {{\hat{Q}}^{k+1}_{\mathsf{a}_1}} \leftarrow  \argmin_Q \alpha \cdot \mathbb{E}_{s \sim \mathcal{D}_{\mathsf{a}_2}, a \sim \pi_{\mathsf{a}_1}}[Q(s, a)]+ \frac{1}{2} \mathbb{E}_{s,a \mathcal{D}_{\mathsf{a}_2}}\left[(Q(s, a) - \mathcal{B}^*{\hat{Q}^{k}_{\mathsf{a}_1}}(s, a) )^2\right]
\end{equation}

Then the resulting Q-function $\hat{Q}_{\mathsf{a}_2} := \lim_{k \rightarrow \infty} \hat{Q}_{\mathsf{a}_1}$, lower bounds $\hat{Q}_{\mathsf{a}_2}$ at all $(s, a)$.

\end{proposition}

\begin{proof}

By setting the derivative of Equation \ref{qiterate} to $0$, we obtain the following expression 

\begin{equation}
    \forall s, a \in \mathcal{D}_{\mathsf{a}_2},k,\quad {{\hat{Q}}^{k+1}_{\mathsf{a}_1}}(s, a) = \mathcal{B}^*{\hat{Q}^{k}_{\mathsf{a}_1}}(s, a) - \alpha \frac{\pi_{\mathsf{a}_1}(a \vert s)}{\pi_{\mathsf{a}_2}(a \vert s)}
\end{equation}

Since $\pi_{\mathsf{a}_1}(a \vert s) > 0, \pi_{\mathsf{a}_2}(a \vert s) > 0, \alpha > 0 $, then at each iteration the Q-value iterate will be underestimated, i.e. ${\hat{Q}}^{k+1}_{\mathsf{a}_1}(s, a) \leq \mathcal{B}^*{\hat{Q}^{k}_{\mathsf{a}_1}}(s, a)$. This underestimation demonstrates that naively using experience realised from another agent may not be appropriate without adjustment. This observation can be extended across all pairwise agent combinations when leveraging different agent experiences. 
\end{proof}

One possible adjustment is to interpret the correction through the lens of value loss function. The interpretation being if the value loss estimation caused under sampling another agent's experience is high, then this is an out of distribution sample which requires higher weight. This provides another interpretation to the {\em Shared Experience Actor-Critic} framework \cite{seac}, whereby the value loss component is a {\em regularization penalty} to the MARL policy. This extends Equation \ref{eq:cqlr} to the multi-agent scenario. We denote $\hat{Q}^{k+1}_{\mathsf{a}_i,\mathsf{a}_j}$ to indicate an update step for agent $\mathsf{a}_i$ using a sample draw from agent $\mathsf{a}_j$ and is shown below with the regularizer component in red:


\begin{align}
{\hat{Q}}^{k+1}_{\mathsf{a}_i,\mathsf{a}_j} &\leftarrow \text{ } { \argmin_Q \alpha \cdot \left( \mathbb{E}_{s \sim \mathcal{D}_{\mathsf{a}_j}, a \sim \pi_{\mathsf{a}_i}}[Q(s, a)] - \mathbb{E}_{s \sim \mathcal{D}_{\mathsf{a}_j}, a \sim \hat{\pi}_{\mathsf{a}_j}(a \vert s) }[Q(s, a)] \right) } \nonumber \\
&+ \frac{1}{2} \mathbb{E}_{s,a \sim \mathcal{D}_{\mathsf{a}_j}}\left[(Q(s, a) - \mathcal{B}^*{\hat{Q}^{k}_{\mathsf{a}_i}}(s, a) )^2\right] +{\color{red} \lambda \cdot\mathbb{E}_{s,\cdot,r, s^\prime \sim \mathcal{D}_{\mathsf{a}_j}, a \sim \pi_{\mathsf{a}_i}} \frac{\pi_{\mathsf{a}_i}(a \vert s)}{\pi_{\mathsf{a}_j}(a \vert s)} \lVert V_{\mathsf{a}_i}(s) - y_{\mathsf{a}_i,\mathsf{a}_j}\rVert }  \nonumber \\
y_{\mathsf{a}_i,\mathsf{a}_j} &= r + \gamma V_{\mathsf{a}_i}(s^\prime), \text{ where } \langle \cdot, \cdot, \cdot, s^\prime \rangle \sim \mathcal{D}_{\mathsf{a}_j} \label{eq:icql}
\end{align}

where $\lambda$ is the hyperparameter which weights the experience of other agents. This hyperparameter is largely insensitive to different values, hence we set the parameter to be $\lambda = 1$ for our experiments. An ablation study over $\lambda$ was conducted in Section \ref{se_sensitivity}. This approach differs from the DQN set up used by \cite{seac} as our approach uses distributional RL as its independent learning algorithm which allows for this correction to be used, which brings the performance in line with the observations made in the {\em Shared Experience Actor Critic} algorithm.

\subsection{Explicit KL-Divergence Regularization through Adaptive Cross-Entropy}

\label{kldiv}

Rather than using experiences from other agents, another approach is to explicitly regularize using KL divergence to encourage structured exploration. This approach was explored as part of Conservative Q-learning \cite{cql} in single-agent framework and {\em Learning Implicit Credit Assignment} (LICA) within the multi-agent framework, however the structured exploration was only through the entropy by treating each agent independently. The {\em adaptive entropy regularization}, $\mathcal{H}$ is created through dynamically controlling the magnitudes of the entropy gradients so that they are inversely proportional to the policy entropy during training

\begin{align*}
    \mathcal{H}_i &:= \alpha H(\pi_{\mathsf{a}_i}) \\
    \partial \mathcal{H}_i &:= - \alpha \cdot \frac{\log \pi_{\mathsf{a}_i} + 1}{H(\pi_{\mathsf{a}_i})}
\end{align*}

where $H$ is the entropy, $\pi_{\mathsf{a}_i}$ is the policy of agent $\mathsf{a}_i$, and $\alpha$ is a constant controlling the regularization strength. In this particularly instance, as we are not operating in an actor critic framework, we are regularizing in maximum entropy RL framework using distributional RL. This approach is used to ensure there is sufficient sustained exploration through training. To extend this to our approach where there is no mixing network or explicit cooperation, we use cross-entropy penalty to encourage agents to explore diverse trajectories.

In order to address this, we first observe the definition of cross entropy $H(p, q) = H(p) + D_{\text{KL}}(p \Vert q)$. Then we construct Q-learning update based on Equation \ref{eq:cqlr} with the pairwise adaptive cross entropy regularization penalty (in red) to be:

\begin{align}
    {{\hat{Q}}^{k+1}_{\mathsf{a}_i}} \leftarrow  \argmin_Q \alpha \cdot \Big(\mathbb{E}_{s \sim {\mathcal{D}_{\mathsf{a}_i}}, a \sim \pi_{\mathsf{a}_i}(a \vert s)}[Q(s, a)] {- \mathbb{E}_{s \sim \mathcal{D}_{\mathsf{a}_i}, a \sim \hat{\pi}_{\mathsf{a}_i}(a \vert s)} [Q(s, a)]}\Big) \nonumber \\
    \label{eq:cqlr}
    + \frac{1}{2} \mathbb{E}_{s,a, r s^\prime  \sim \mathcal{D}_{\mathsf{a}_i}}\left[(Q(s, a) - \mathcal{B}^*{\hat{Q}^{k}}(s, a) )^2\right] + {\color{red} \lambda \sum_{\mathsf{a}_j \in \mathcal{N}}\left( H(\pi_{\mathsf{a}_i}) + D_{\text{KL}} (\pi_{\mathsf{a}_i} \Vert \pi_{\mathsf{a}_j})  \right)}
\end{align}

Where $\lambda$ is the regularization strength. Under this framework, in addition to the adaptive regularizer used in LICA, the regularizer would leverage pairwise experience information. Since cross-entropy is defined as $H(p, q) = H(p) + D_{\text{KL}}(p \Vert q)$, then the non-negativity of the KL divergence on the right-hand side implies that a agent policies which behave in the same manner induces a regularizer penalty, encouraging diversity, which is offset by entropy. Not only does our approach extend the LICA adaptive entropy, it is a natural extension to a MARL variation of CQL models introduced in Section \ref{cqlr}.

\section{Experiments}

\begin{figure}[ht]
\centering
\begin{tabular}{cccc}
\includegraphics[width=0.2\textwidth,height=0.2\textwidth]{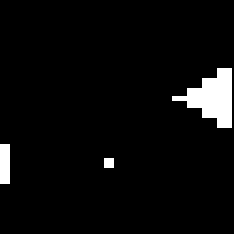} & \includegraphics[width=0.2\textwidth,height=0.2\textwidth]{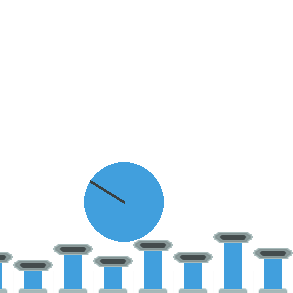} &
\includegraphics[width=0.2\textwidth,height=0.2\textwidth]{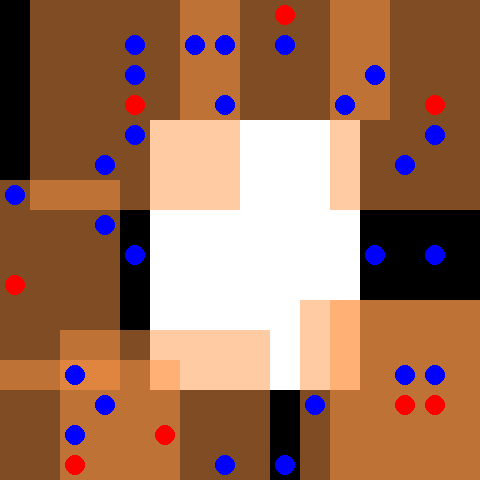} &
\includegraphics[width=0.2\textwidth,height=0.2\textwidth]{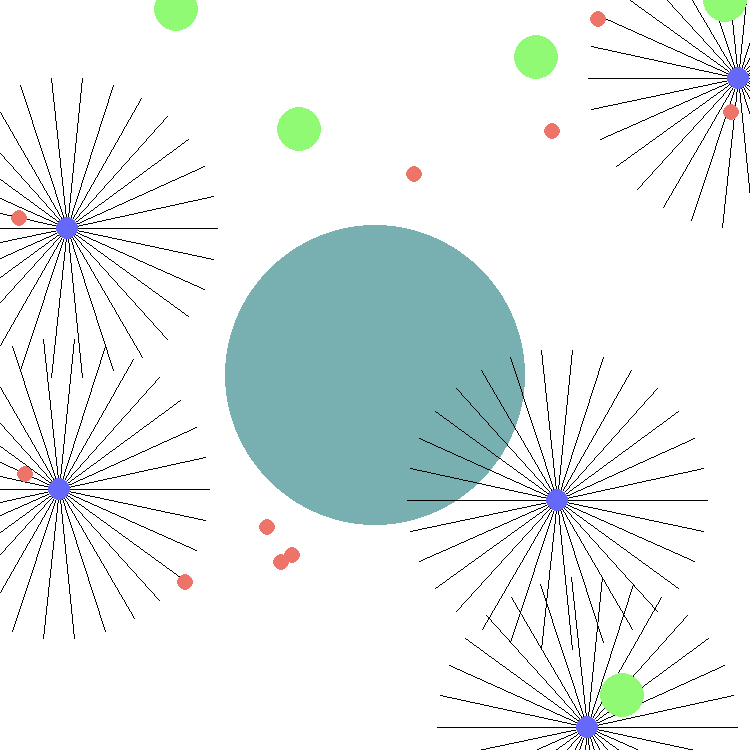}
\end{tabular}
\caption{Left to right: Pong, Pistonball, Pursuit, Waterworld environments.}
\label{fig:pettingzoo}
\end{figure}

In this section we describe experiments we use to empirically justify our algorithm and approach. We use several reinforcement learning benchmarks \cite{terry2020pettingzoo,maddpg,gupta2017cooperative}, shown in Figure \ref{fig:pettingzoo}.

{\em Cooperative Pong} is a multi-agent game of pong from ``Butterfly'' environments \cite{terry2020pettingzoo}, where the goal is for both agents (paddles) to keep the ball in play. The game is over if the ball goes out of bounds from either the left or right edge of the screen. In this setting the action space is discrete, and the paddles can either move up or down. To make learning more challenging the right paddle is tiered cake-shaped by default, and the observations of each agent are its own half of the screen. The agents receive a positive, fixed, combined reward based on successful completion of a game frame, otherwise a negative reward is given, and the game ends. 

{\em Pistonball} is a physics based cooperative game from ``Butterfly'' environments \cite{terry2020pettingzoo} where the goal is move the a ball to the left wall of the game border by activating vertically moving pistons. Each piston agent's observation is an RGB image of is local surrounding area. The agents have a discrete action space, and can choose to stay still, move down or move up. Pistons must learn highly coordinated emergent behavior to achieve an optimal policy for the environment. Each agent gets a reward based on the overall movement of the ball, and how much the ball moved in a left-wards direction if it was in the agent's local observation. 

{\em Pursuit} is an environment from the ``SISL'' set of environments \cite{gupta2017cooperative}, where our agents are control of pursuer agents, which aim to capture randomly controlled evader agents. Evader agents are only removed from the environment when pursuers fully surround an evader. Agents receive a small reward when they touch an evader, and a large reward when they successfully surround and remove an evader agent. In this scenario, each pursuer agent observes a local RGB grid, and operate in a discrete action space. 

{\em Waterworld} is a simulation of archea trying to navigate and surviving their environment from the ``SISL'' set of environments \cite{gupta2017cooperative}. The goal of the agent is to control the archea to simultaneously avoid poison and find food. The original environment consists of a continuous action space, indicating their horizontal and vertical thrust within the environment. We instead modify this to a discrete action space, in order to assess Q-learning approaches. Within this environment, the agent's observation is based on range limited sensors to detect neighboring entities, indicated by black lines in the right-most image in Figure \ref{fig:pettingzoo}. 

{\em Reference} and {\em Spread} are environments consisting of agents and landmarks from the ``MPE'' set of environments \cite{maddpg}. In {\em Reference}, each agent wants to get closer to their target landmark, which is known only by other agents. The agents are rewarded by the distance to their target landmark. In {\em Spread} Agents must learn to cover each landmark without colliding with other agents. The agents are rewarded based on how far the closest agent is to each landmark, and are penalized if they collide with other agents. For both environments, the agent's observation space consists of distance to every landmark, and the action space consists of the movement action. In addition the {\em Reference} environment has an additional actions to communicate with other agents.

\subsection{Architecture Overview and Implementation Details}

Across all environments, we consistently preprocessed the inputs as suggested in the PettingZoo benchmarks \cite{terry2020pettingzoo}. We evaluated on 1000 evaluation episode rollouts (separate from the train distributions) every training iteration and used the average score and variance for the plots and tables, shown in Figure \ref{resall}

\subsection{Hyper-parameters}
\begin{table*}[!htb]
\centering
\begin{tabular}{c|c}
\hline
\textbf{Hyperparameters}    & \textbf{Value} \\ \hline
Policy Hidden Sizes         & {[}64, 64, 64{]} \\
Mixer Hidden Sizes         & {[}32, 32, 32{]} \\
Policy Hidden Activation    & ReLU           \\
Target Network $\tau$          & 0.005          \\
Learning Rate               & 0.0003         \\
Batch Size                  & 256            \\
Replay Buffer Size          & 1000000        \\
Number of pretraining steps & 1000           \\
Steps per Iteration         & 1000           \\
Discount                    & 0.99           \\
Reward Scale                & 1              \\ \hline
\end{tabular}
\caption{Hyper-parameters used for MARL experiments}
\label{app:tab:hyper}
\end{table*}
\label{ref:hyper}

The policies leverage the default hyper-parameters based on the official implementation of LICA, QMIX, IQL and their accompanying code based on the pymarl library \cite{qmix,lica,qtran}. SEAC was re-implemented in rlkit framework with the same hyperparameters proposed in their paper, but using Soft Actor-Critic instead of A2C. Similarly, MADDPG extended the rlkit implementation of DDPG to support multi-agent setting. In order to facilitate discrete action spaces gumbel softmax \cite{gumbel_softmax1,gumbel_softmax2} is used as proposed by the approach taken by by SEAC \cite{seac}. Similar to the usage in the pymarl library, parameter sharing was used in all models in order to improve the general performance of the agents.

\subsection{Results}

\begin{figure}[!htb]
  \begin{center}
\begin{tabular}{c}
\includegraphics[width=0.75\textwidth]{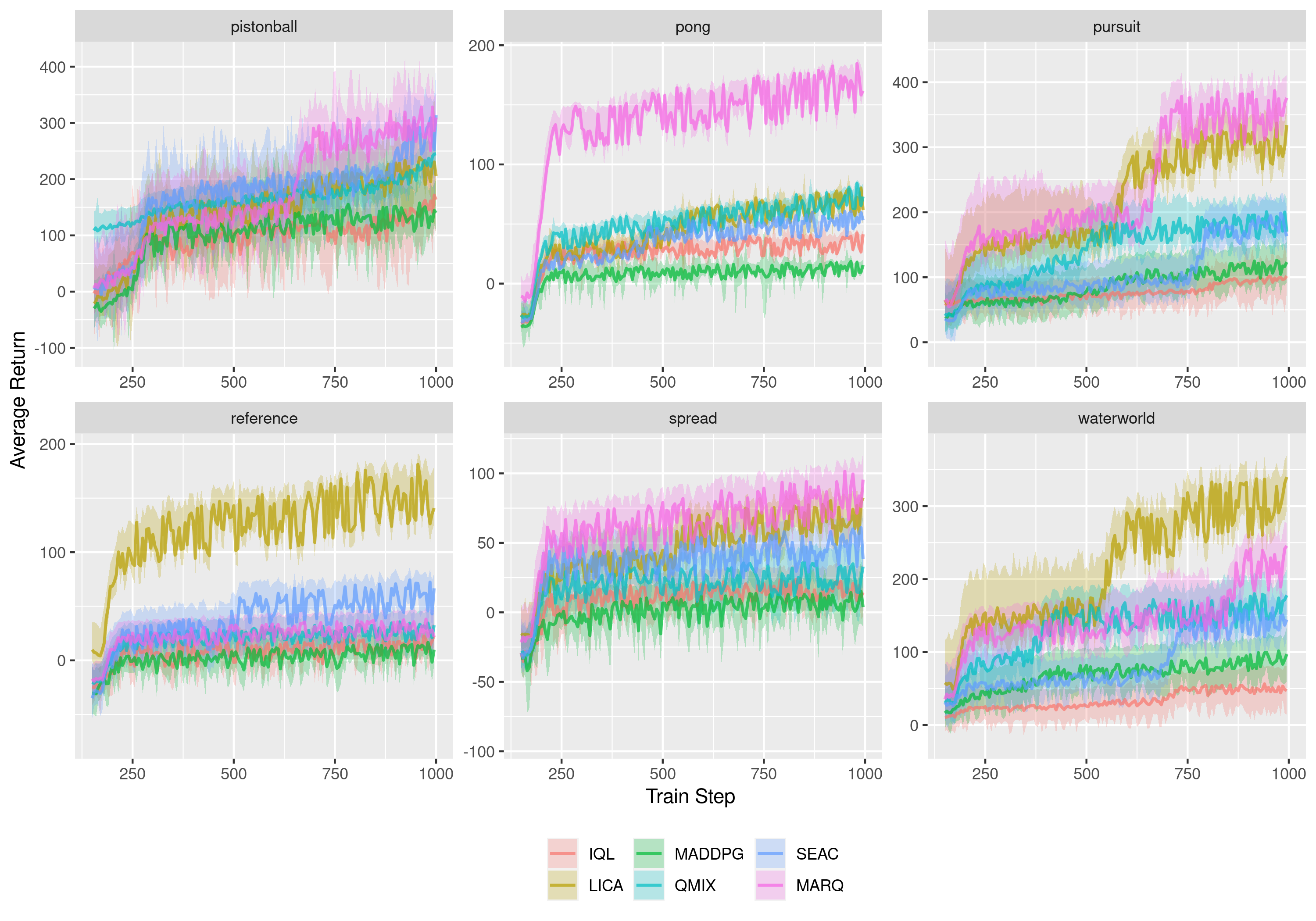} \\
\includegraphics[width=0.75\textwidth]{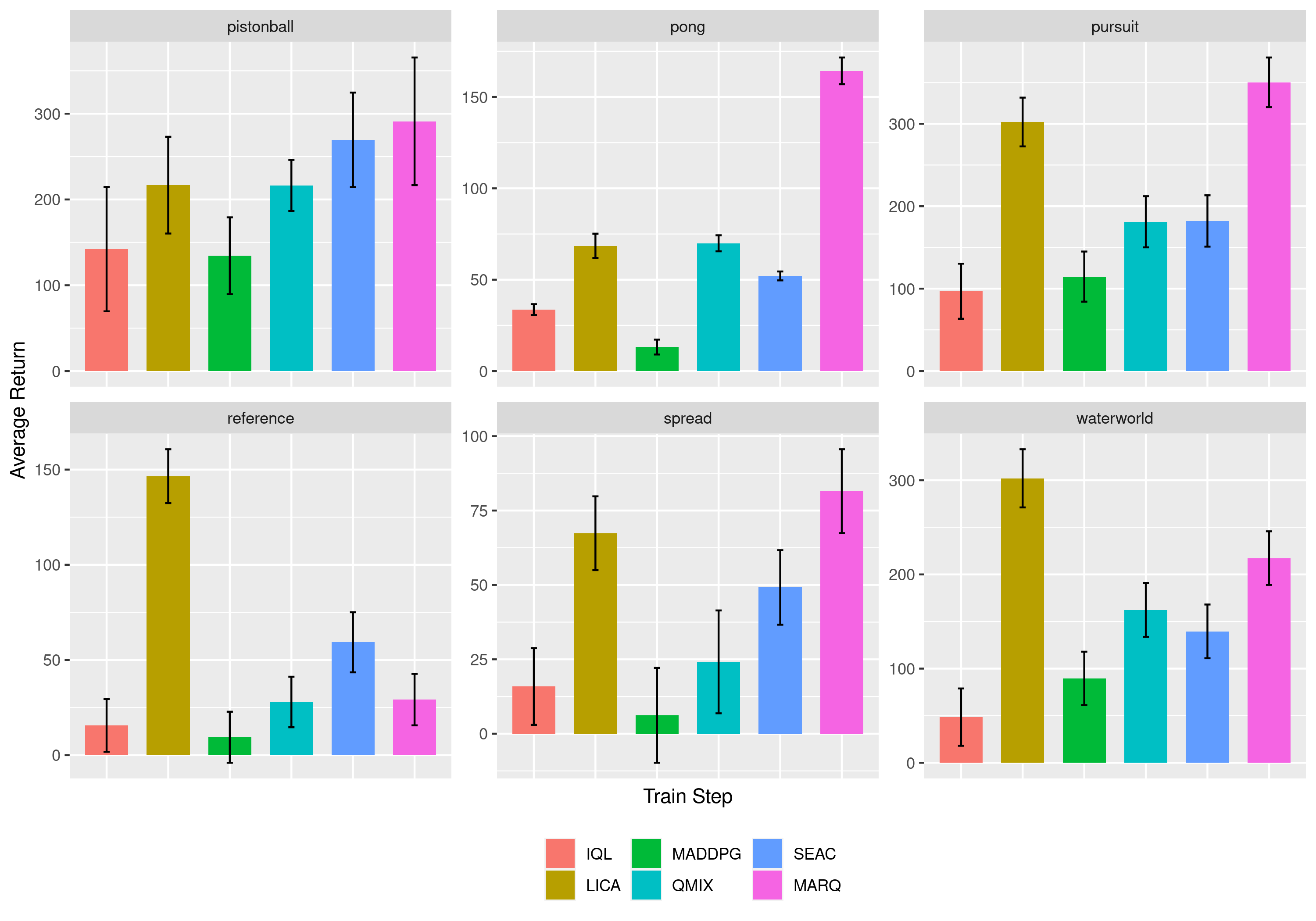}
\end{tabular}
  \end{center}
\caption{Performance over a variety of benchmark environments. The bar graph report the mean over the last 10 evaluation steps, where each evaluation point is average over 1000 episodes. Error bars for plots represent the standard deviation over these rollouts.}
\label{resall}
\end{figure}

We demonstrate that our approach MARQ is comparable to existing state of the art approaches, and performs well against approaches which use explicit cooperation as shown in Figure \ref{resall}. 

\begin{figure}[!htb]
  \begin{center}
    \includegraphics[width=0.95\textwidth]{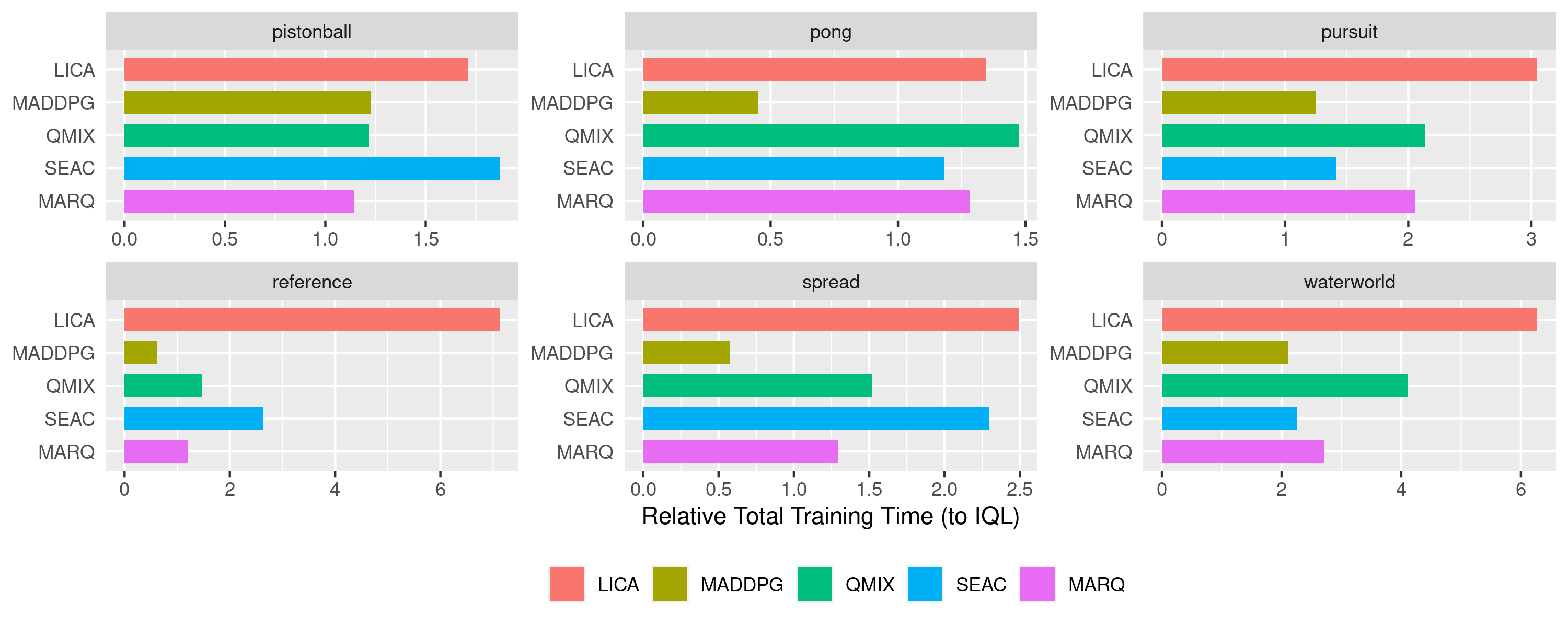}
  \end{center}
\caption{Comparison of the relative total time taken in training the different approaches, using IQL as the benchmark.}
\label{restime}
\end{figure}

From the plots, we observe our approach competitive with the state-of-the-art approaches across a range of benchmark environments as shown in Figure \ref{resall} and generally completes training in less time compared with approaches which use a mixer network. There are some performance improvements over actor critic methods (SEAC) as it does not need to train both an actor and critic network as shown in Figure \ref{restime}.

The increase in training time compared with IQL can be explained through the difference in training distribution RL model in MARQ compared with vanilla DQN in the IQL implementation.

\begin{figure}[!htb]
\centering
  \begin{center}
\begin{tabular}{c}
\includegraphics[width=0.75\textwidth]{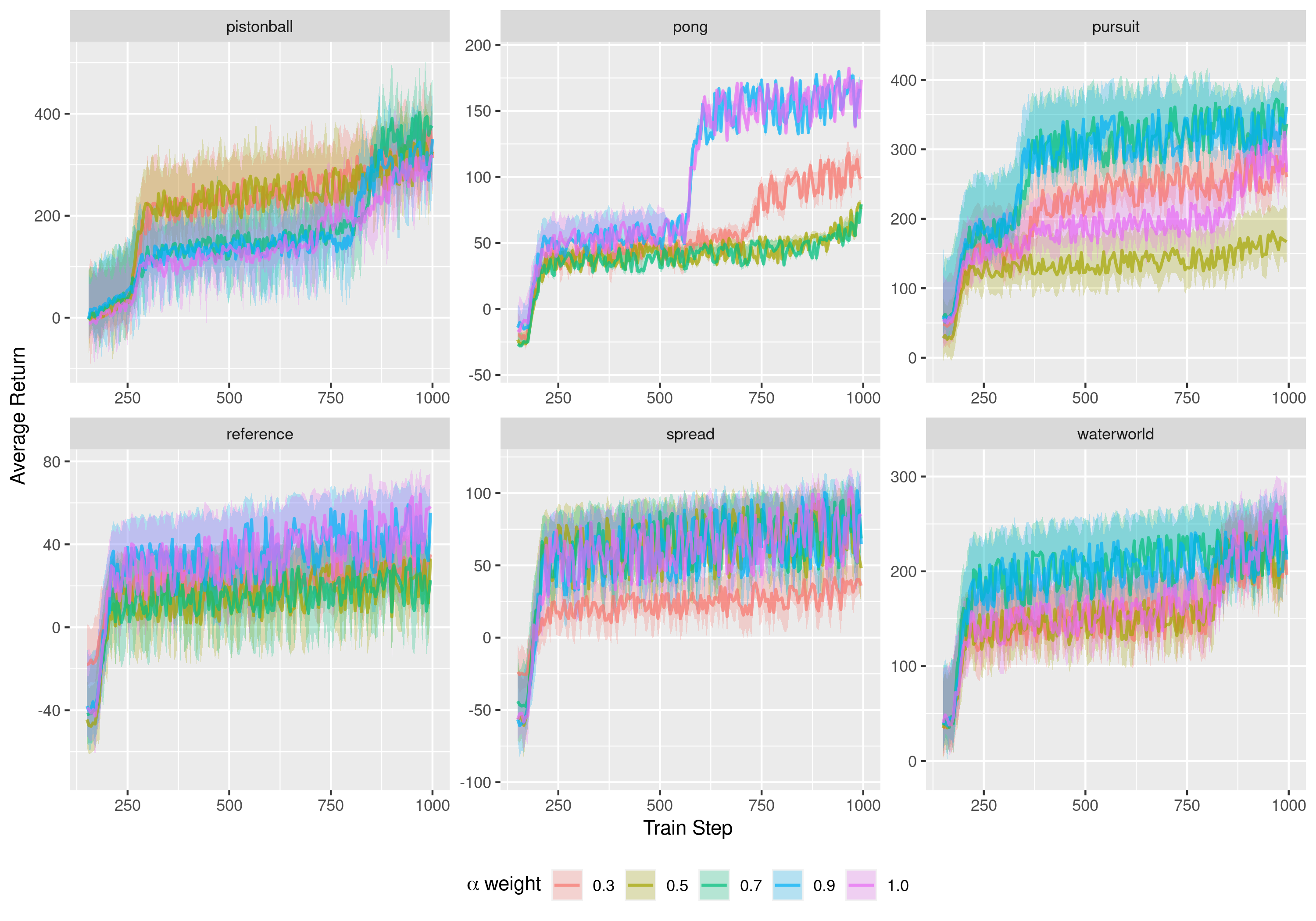}  \\
\includegraphics[width=0.75\textwidth]{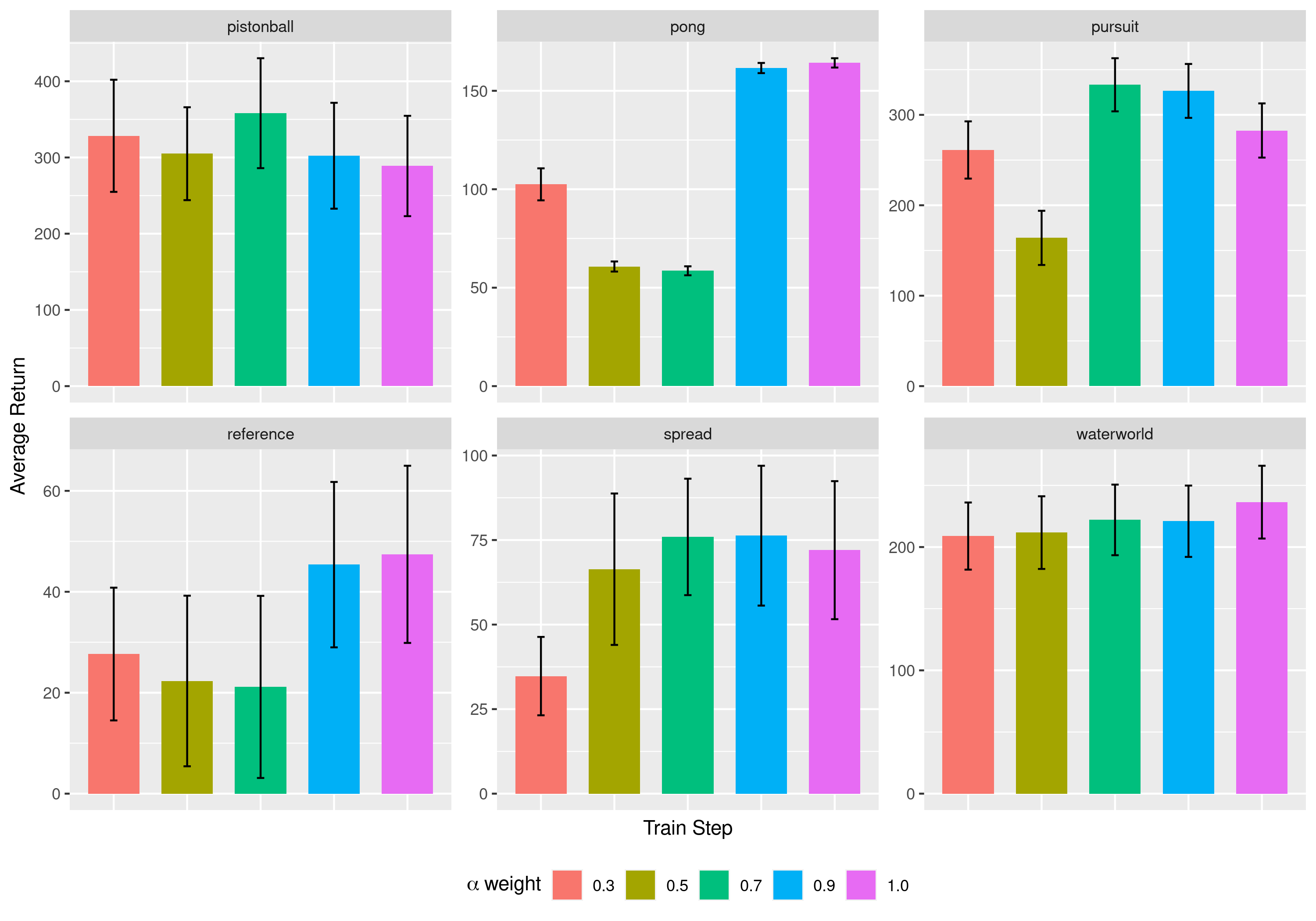}
\end{tabular}
  \end{center}
\caption{Comparison the sensitivity of the $\lambda$ parameter when correcting for shared experiences as described in Section \ref{sharedexp}. The bar graph report the mean over the last 10 evaluation steps, where each evaluation point is average over 1000 episodes. Error bars for plots represent the standard deviation over these rollouts.}
\label{plot:abl}
\end{figure}

\begin{figure}[!htb]
\centering
  \begin{center}
\begin{tabular}{c}
\includegraphics[width=0.75\textwidth]{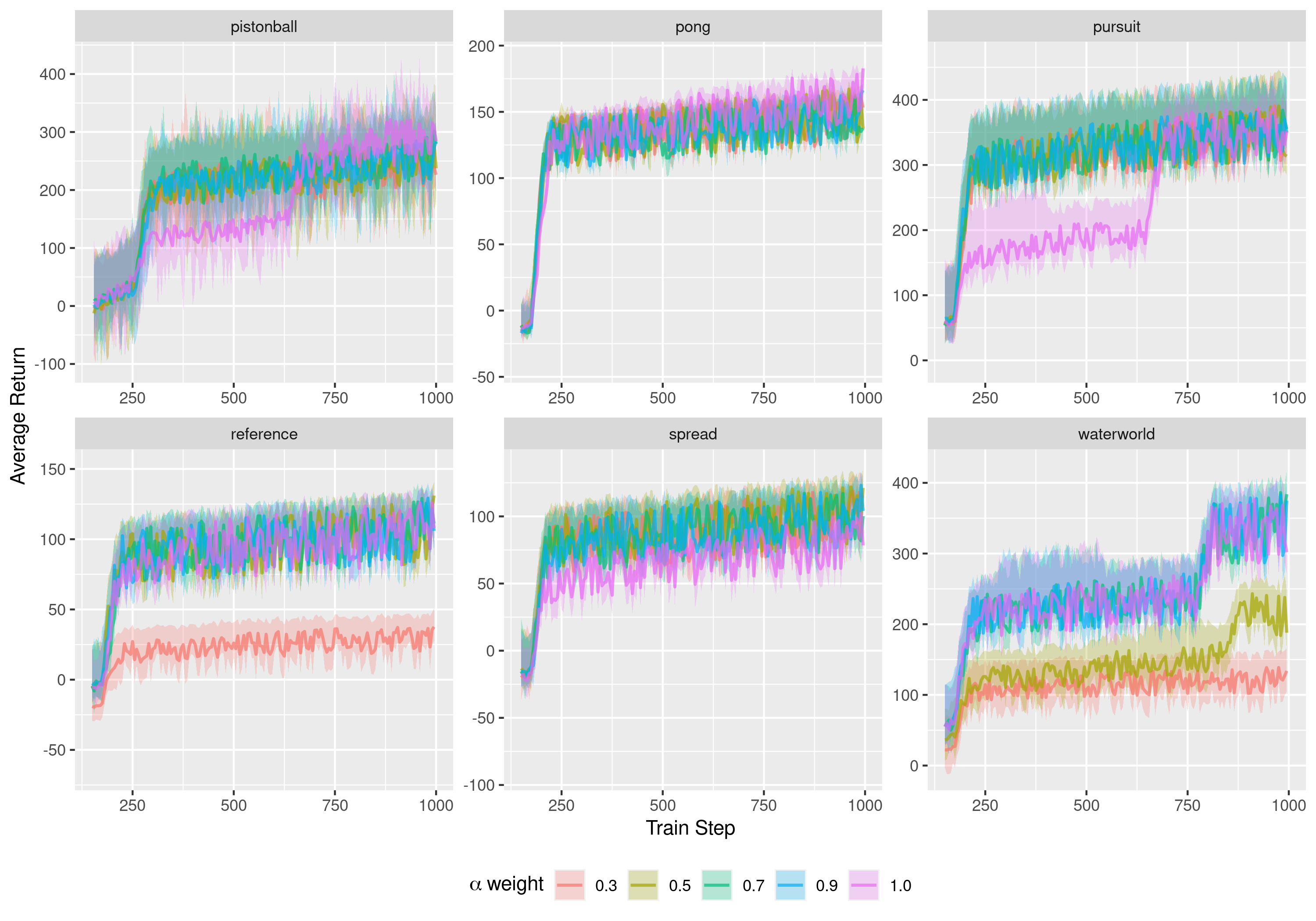}  \\
\includegraphics[width=0.75\textwidth]{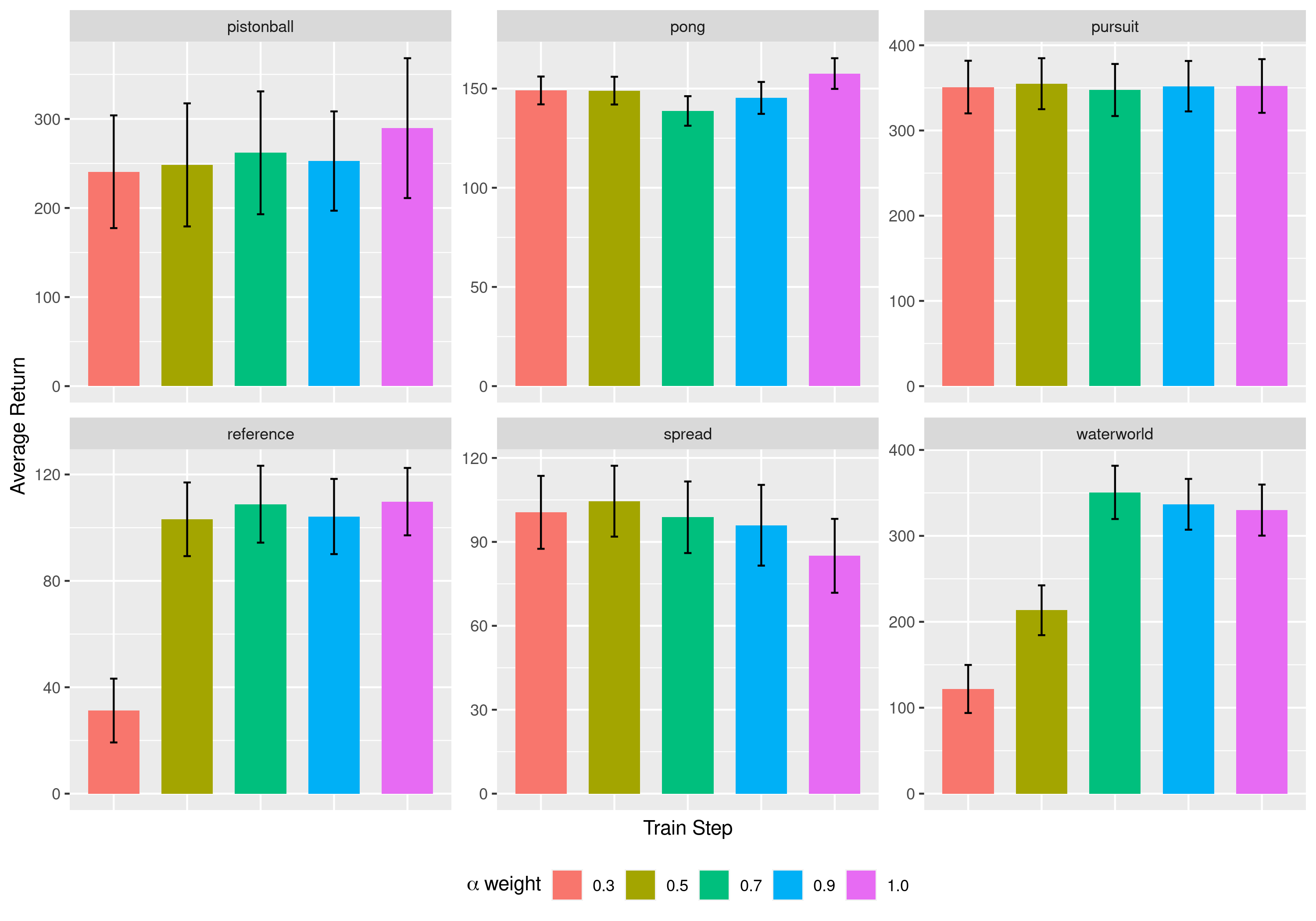}
\end{tabular}
  \end{center}
\caption{Comparison the sensitivity of the $\lambda$ parameter when correcting for KL-divergence variation as described in Section \ref{kldiv}. The bar graph report the mean over the last 10 evaluation steps, where each evaluation point is average over 1000 episodes. Error bars for plots represent the standard deviation over these rollouts.}
\label{plot:ablkl}
\end{figure}

\subsection{Ablation}

\label{se_sensitivity}

We compare our results with different hyperparameters for our regularizer. We demonstrate empirically for both cases, regularization in the shared experience framework and regularization using KL-divergence, our algorithm is not sensitive to weights. As expected, when $lambda \rightarrow 0$, the model approaches the unregularized variation, which leads to decreased performance. This is shown for Shared Experience variation in Figure \ref{plot:abl} and for KL-divergence variation in Figure \ref{plot:ablkl}. In conclusion MARQ is not sensitive to the choice of hyperparameter relating to the level of shared experience or divergence weight and emits a stable policy.


\section{Conclusion}

We have demonstrated the efficacy of our approach through demonstrated the speed of convergence and also the overall performance against other state-of-the-art algorithms and have demonstrated the algorithms ability to generalise. Furthermore we have justified the construction of our algorithm through ablation studies, revealing the stability of our approach. 

Future work could include an automated way of tuning the amount of sharing dependent on the type of multi-agent environment used. For example, training MARL algorithms in a competitive multi-agent environment in the implicit importance sampling perform worse compared with an orthogonal search strategy in adaptive cross-entropy which promotes more diverse behavior for learning. The conservative Q-learning bound assists in understanding and optimising agents when the joint action space from a global perspective is sparse. 

\clearpage

\bibliography{acml20}
\bibliographystyle{plain}

\end{document}